\documentclass{article}

\usepackage[dblblindworkshop,final]{neurips_2026}
\workshoptitle{Developmental Perspectives on AI}

\usepackage[utf8]{inputenc}
\usepackage[T1]{fontenc}
\usepackage{hyperref}
\usepackage{url}
\usepackage{booktabs}
\usepackage{amsfonts}
\usepackage{nicefrac}
\usepackage{microtype}
\usepackage{xcolor}
\usepackage{graphicx}
\usepackage{amsmath} 

\title{Attention Function as an Intrinsic Inductive Bias: How Models' Behavior Diverges in Novel Contexts}

\author{%
  Dong Gyun Kang$^{1,2}$ \qquad
  Megha Thukral$^{1}$ \qquad
  Kwangsoo Kim$^{2}$ \\[6pt]
  $^{1}$College of Computing, Georgia Institute of Technology \\
  $^{2}$Department of Transdisciplinary Medicine, Seoul National University Hospital \\[6pt]
  \texttt{dkang335@gatech.edu} \quad
  \texttt{mthukral3@gatech.edu} \quad
  \texttt{kwangsookim@snu.ac.kr}
}

\begin{document}

\maketitle

\begin{abstract}
Developmental psychology holds that certain priors are given to infants prior to experience rather than induced from data, and that the influence of such priors is suppressed under strong, well-constrained conditions but reasserts itself under weak ones. 
We ask whether an analogous principle holds for the Transformer: can the activation function given to attention heads serve as an intrinsic inductive bias? 
We propose \textbf{Mixture of Function Attention (MoFA)}, a parameter-free modification to multi-head attention that fixes a ratio of softmax and sigmoid heads before training.
Across five ratios, a 124M-parameter GPT-2 model, and five seeds, we find that this given ratio has little effect in-distribution---differences between ratios are statistically negligible for moderate mixtures and remain small even at the extremes---but its influence re-emerges sharply under zero-shot distribution shift across 15 out-of-distribution domains.
Perplexity gaps between ratios widen by more than an order of magnitude on several domains, and the best-performing ratio tracks a single axis of domain structure, separating short, informal text (softmax-favoring) from technical, long-form text (sigmoid-favoring), that explains $78.3\%$ of the variance in domain response. 
This reorganization is visible at the head level: 
sigmoid heads show an accelerating drop in attention entropy as their ratio increases, while softmax heads respond more modestly, yielding a consistent division of labor between the two head types. 
Our results suggest that activation choice functions as a given prior whose influence is masked in-distribution and re-emerges out-of-distribution.
\end{abstract}

\section{Introduction}
\label{sec:intro}
Developmental psychology posits that infants are not blank slates: certain domains—objects, number, space, agency—are structured by core knowledge systems that are present prior to experience~\cite{spelke2007core}. These systems function as given priors, shaping how new information is interpreted from the earliest stages of learning. We take inspiration from this view and ask an analogous question for the Transformer~\cite{vaswani2017attention}: can the attention function given to a network serve as an intrinsic inductive bias, shaping how it organizes its representations—and how it generalizes to novel contexts? Akin to how infants generalize learned associations to new environments despite limited experience~\cite{kangas2011transfer}.

We hypothesize that in-distribution training constitutes a well-constrained regime, and can constrain model behavior, potentially masking differences in the inductive biases imposed by different attention ratios. In zero-shot out-of-distribution (OOD) settings, these constraints are weaker because the model receives no task-specific adaptation to the new distribution, allowing differences induced by the attention ratio to become more apparent.
This pattern offers a parallel to a core finding in Bayesian cognitive science: a given prior is largely masked by strong, well-constrained sensory evidence, but reasserts its influence on inference once that evidence becomes weak or ambiguous~\cite{knill2004bayesian}.

The Transformer has applied the same activation function across all attention heads since its inception; even attempts to replace softmax altogether have kept the substitute uniform across the entire model~\citep{Wortsman2023ReplacingSW, saratchandran2025polynomial}—the given prior has always been uniform. We ask whether this prior can instead be given as a mixture, and propose \textbf{Mixture of Function Attention (MoFA)}, which assigns different activation functions to different heads as a head-level decision. We instantiate \emph{MoFA} with softmax and sigmoid: sigmoid is an appealing partner since \citet{ramapuram2025sigmoid} established a principled normalization method for it, and its output scale is already comparable to softmax's, requiring little extra normalization to combine the two within the same layer. Beyond this practical fit, the two are functionally distinct: softmax competes keys against each other via a sum-to-one constraint, while sigmoid gates each independently, allowing dense, distributed attention. We evaluate \emph{MoFA} across a range of sigmoid-to-softmax ratios, testing zero-shot generalization across a broad suite of out-of-distribution domains.

Our contributions are as follows:
\begin{itemize}
\item We propose \emph{MoFA}, a parameter-free modification to Multi-Head Attention that mixes activation functions across heads at a fixed ratio, requiring no architectural changes beyond the choice of activation.
\item We demonstrate empirically across 15 OOD domains that no single activation function is universally optimal, and that the best-performing ratio depends on the alignment between its inductive bias and the structure of the OOD domain.
\item We show that mixed attention head models exhibit a division of labor at the head level, with sigmoid and softmax heads operating at consistently different attention entropy and adopting complementary roles.
\item We frame the given ratio of attention functions as an intrinsic prior inspired by developmental and cognitive science, and show that its influence is suppressed in-distribution but re-emerges under distribution shifts.
\end{itemize}

\section{Related Work}
\label{sec:relwork}
\paragraph{Developmentally Inspired Architectures.}
\citet{lattice2023} infuse lattice symmetry priors into attention mechanisms to improve sample efficiency on abstract geometric reasoning, directly building architectural constraints from a Core Knowledge domain (space and geometry) into the attention operation.
Similarly, \citet{chakravarthy2023spotlight} incorporate a spatial-locality prior into object-centric vision models, motivated by the observation that human visual attention is far more spatially constrained than the diffuse competition used in standard slot-based architectures.
Both share our premise that such biases are best given, not induced, though each ties it to a hard architectural constraint on a specific domain.

\paragraph{Modifying Softmax Attention.}
The dominance of softmax in attention mechanisms has prompted several attempts to replace or modulate it: \citet{katharopoulos2020transformers} approximate softmax via a kernel decomposition for linear-time attention, \citet{ramapuram2025sigmoid} establish sigmoid self-attention as a theoretically grounded and hardware-efficient substitute, and \citet{qiu2025gated} apply a head-specific sigmoid gate after the scaled dot-product output to improve stability and long-context extrapolation. In each case, the activation or gating choice is applied uniformly across all heads, rather than as a per-head decision, as \emph{MoFA} proposes.

\paragraph{Head-Level Heterogeneity.}
The idea that individual heads within the same layer may benefit from different computational structures has recently gained traction.
\citet{tan2025hydrahead} propose HydraHead, which hybridizes full attention and linear attention at the head level, motivated by computational efficiency for long-context processing, with head selection determined by interpretability-based importance scoring.
\citet{zhang2024moh} treat attention heads as experts in a Mixture-of-Experts framework, routing tokens to a dynamic subset of heads at inference time.
Both approaches vary the \emph{mechanism} across heads; neither addresses the question of activation function diversity within the same dot-product attention structure, which we introduce here.

\paragraph{Attention Head as Kernel.}
A growing body of work reinterprets individual attention heads as kernel machines, providing theoretical grounding for treating activation choice as a kernel choice.
\citet{tsai2019transformer} first cast dot-product attention as a kernel smoother, where softmax's normalization corresponds to one particular non-negative kernel among a much larger space of valid choices; \citet{wright2021transformers} sharpen this view, proving that dot-product attention is exactly the reproducing kernel of a pair of Banach spaces with an infinite-dimensional feature map.
\citet{cheng2024transformers} extend the correspondence to the learning algorithm a head implements in-context, showing that when an attention head's non-linearity matches a kernel, the head performs functional gradient descent in the space that kernel induces during the forward pass itself.
This view directly motivates \emph{MoFA}: if activation choice fixes the function space a head can search, then mixing activations across heads within a layer amounts to searching multiple function spaces in parallel, rather than committing the entire model to one.

\section{Method}
\label{sec:method}
\subsection{Preliminaries}

Standard scaled dot-product attention computes, for a sequence of length $T$
with queries $Q$, keys $K$, and values $V \in \mathbb{R}^{T \times d_k}$:
\begin{equation}
    \text{Attn}(Q, K, V) = f\!\left(\frac{QK^\top}{\sqrt{d_k}}\right) V,
\end{equation}
where $f = \text{softmax}$ is applied row-wise, producing a probability
distribution over positions. In multi-head attention (MHA), this operation is
repeated across $H$ heads operating on independent low-dimensional projections,
with outputs concatenated and projected back.

\subsection{Mixture of Function Attention}
In contrast to previous attempts at replacing softmax outright, MoFA gives the activation choice as a fixed prior: the ratio of activation functions across heads is fixed before training and never learned. Concretely, MoFA partitions the $H$ heads into two groups of sizes $H_{sfx} = i$ and $H_{sig} = j$
($i + j = H$), assigning a distinct scoring activation to each:
\begin{align}
    a^{(p)}_{\text{sfx}} &= \text{softmax}\!\left(
        \frac{Q^{(p)} {K^{(p)}}^\top}{\sqrt{d_k}} + M_{\text{causal}}
    \right), \quad p = 1, \ldots, i, \\[6pt]
    a^{(q)}_{\text{sig}} &= \sigma\!\left(
        \frac{Q^{(q)} {K^{(q)}}^\top}{\sqrt{d_k}} - \log t + M_{\text{causal}}
    \right), \quad q = 1, \ldots, j,
\end{align}
where $M_{\text{causal}}$ is the causal mask (setting future positions to
$-\infty$), $\sigma$ denotes the sigmoid function, and $t$ is the causal window
size at each query position (the number of tokens the query can attend to).
The $-\log t$ shift normalises the expected magnitude of sigmoid outputs to be
comparable across sequence positions, analogous to the row-sum normalisation
implicit in softmax; this stabilisation technique follows \citet{ramapuram2025sigmoid}.
All heads share a single QKV projection matrix; no additional parameters are
introduced. The two groups produce attention maps with fundamentally different
properties: softmax heads enforce a competitive, zero-sum allocation (rows sum
to 1), while sigmoid heads gate each key position independently, permitting
dense and distributed attention patterns. The final output is:
\begin{equation}
    \text{MoFA}(X) = W_O \, \text{concat}\!\left(
        a^{(1)}_{\text{sfx}} V^{(1)},\ \ldots,\
        a^{(i)}_{\text{sfx}} V^{(i)},\
        a^{(1)}_{\text{sig}} V^{(1)},\ \ldots,\
        a^{(j)}_{\text{sig}} V^{(j)}
    \right).
\end{equation}

Head assignment follows a fixed, contiguous partition by index: within
each layer, the first $H_{\text{sfx}}$ heads are assigned softmax and the
remaining $H_{\text{sig}}$ heads are assigned sigmoid, identically across
all layers and fixed prior to training. This partition determines,
before any exposure to data, which combination of function spaces the
model can draw on—a structural constraint fixed in advance, much like a
given prior, rather than one learned or adapted during training.

\section{Experiments}
\label{sec:experiment}
\subsection{Setup}

\paragraph{Model.}
We train GPT-2-style decoder-only language models~\cite{radford2019language}
with 124M parameters (12 layers, 12 heads, embedding dimension 768, context length 1024).
A custom BPE tokenizer with a vocabulary of 50,000 tokens is trained on OpenWebText, so that all model components—including the tokenizer—are learned from scratch under identical conditions. 
All models use pre-LayerNorm blocks and weight tying between the token embedding and the language model head.

\paragraph{Training.}
All models are trained for 50,000 steps with a cosine learning rate schedule, peak learning rate $3 \times 10^{-4}$, linear warmup, AdamW optimizer ($\beta_1 = 0.9$, $\beta_2 = 0.95$), and gradient clipping at 1.0.
Validation perplexity is evaluated every 500 steps; the checkpoint with the lowest validation loss is retained.
Each model is trained on a single NVIDIA H200 GPU.

\paragraph{Datasets.}
Our models are pretrained on OpenWebText \citep{gokaslan2019openwebtext}. To evaluate out-of-distribution generalization, we construct a suite of 15 domains spanning code, narrative, news, scientific, encyclopedic, social media, legal, dialogue, and structured-reasoning text, drawn from established benchmarks including \citet{husain2019codesearchnet}, \citet{zhu2015aligning}, \citet{see2017get}, \citet{cohan2018discourse}, \citet{merity2016pointer}, \citet{kim2019abstractive}, \citet{kornilova2019billsum}, \citet{li2017dailydialog}, \citet{rashkin2019towards}, \citet{barbieri2020tweeteval}, \citet{socher2013recursive}, \citet{paperno2016lambada}, \citet{cobbe2021training}, and \citet{marcus1993building}. For each domain we use the corresponding Hugging Face dataset and split. We provide the full dataset identifiers, splitsin Table~\ref{tab:datasets} in Appendix~\ref{app:datasets}.

\paragraph{Controlled comparison.}

All five ratio configurations ($H_{sig} \in \{0, 3, 6, 9, 12\}$) share
identical data order, weight initialization, and training hyperparameters
across five independent runs (seeds 42, 1005, 1111, 2026, 9999), differing
only in the sigmoid head ratio. The softmax-only model ($H_{sig} = 0$) serves
as the baseline.

\subsection{In-Domain Validation.}
\label{sec:in-domain}
\begin{table}[h]
\centering
\caption{Internal validation perplexity on OpenWebText by sigmoid head ratio,
         averaged over 5 seeds ($\pm$ std). Paired $t$-test against the softmax-only
         baseline; $^{*}$~$p < 0.01$.}
\label{tab:internal_ablation}
\begin{tabular}{ccccc}
\toprule
\textbf{Softmax} & \textbf{Sigmoid} & \textbf{Val Loss} & \textbf{Val PPL} & \textbf{\textit{p}} \\
\midrule
12 & 0  & $3.0568 \pm 0.0007$ & $21.26 \pm 0.01$ & --- \\
9  & 3  & $3.0571 \pm 0.0015$ & $21.26 \pm 0.03$ & $0.693$ \\
6  & 6  & $3.0621 \pm 0.0017$ & $21.37 \pm 0.04$ & $0.001^{*}$ \\
3  & 9  & $3.0673 \pm 0.0011$ & $21.49 \pm 0.02$ & $<0.001^{*}$ \\
0  & 12 & $3.0814 \pm 0.0027$ & $21.79 \pm 0.06$ & $<0.001^{*}$ \\
\bottomrule
\end{tabular}
\end{table}

Table~\ref{tab:internal_ablation} reports validation perplexity on a held-out
OpenWebText split. We assess statistical significance via paired $t$-tests
against the baseline across the five seeds.
The configuration with $H_{sig} = 3$ achieves validation perplexity identical
to the baseline ($21.26$, $p = 0.693$), indicating that introducing sigmoid
heads does not degrade in-domain performance when the ratio is moderate.
Beyond $H_{sig} = 3$, perplexity increases consistently and monotonically with
the number of sigmoid heads, reaching a maximum absolute degradation of $0.53$
PPL at $H_{sig} = 12$ ($21.26 \to 21.79$)---a modest effect size, though
statistically significant for $H_{sig} \geq 6$ ($p < 0.01$), consistent with
the intuition that OpenWebText---a general web corpus---\textbf{favors the
competitive, selective attention patterns of softmax.}

\subsection{Out-of-Distribution Evaluation.}

To test how the given ratio's influence generalizes to novel environments, we
evaluate all five ratio configurations zero-shot on 15 domains under
out-of-distribution shift, spanning code, biomedical text, scientific
literature, news, social media, dialogue, and standard language modeling
benchmarks (Table~\ref{tab:ood}). No fine-tuning or domain adaptation is
performed.

\subsubsection{Performance by Domain.}
\label{subsubsec:domain_perfomance}

\begin{figure}[htbp]
    \centering
    \includegraphics[width=\textwidth]{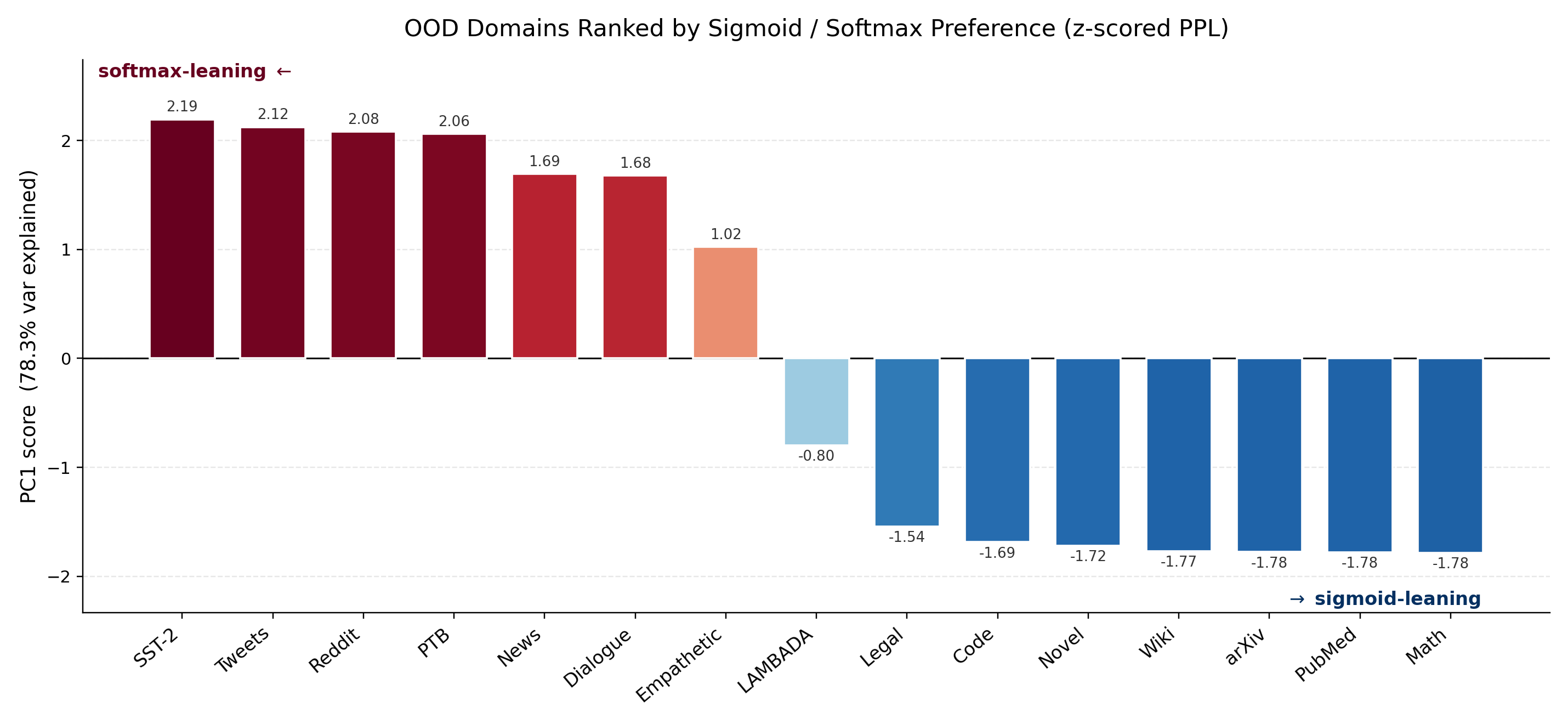}
    \caption{OOD domains ranked by their first principal component (PC1) of
             row-normalized zero-shot perplexity across the five ratios
             ($78.3\%$ variance explained). Domains separate into a
             softmax-leaning group (short, informal text) and a
             sigmoid-leaning group (technical, long-form text); conversational
             domains (Dialogue, Empathetic) sit closest to the boundary.
             Table~\ref{tab:ood} follows this ordering.}
    \label{fig:pca1_domain}
\end{figure}

Table~\ref{tab:ood} reports zero-shot perplexity for all five ratios across
the 15 domains. At a high level, no single ratio dominates across domains:
each of the five configurations is the best choice on at least one domain,
and the two pure endpoints trade off—one excelling where the other falters.
This domain-dependence tracks the axis identified in
Figure~\ref{fig:pca1_domain}: softmax tends to suit short, informal text,
sigmoid tends to suit technical and long-form text, and conversational
domains fall in between. We unpack these patterns, and their exceptions,
below.

\begin{table*}[htbp]
\centering
\caption{Zero-shot perplexity (mean $\pm$ std over 5 seeds) across OOD domains by sigmoid head count.
         Columns denote $H_{\text{sfx}}{:}H_{\text{sig}}$, the number of softmax vs.\ sigmoid heads out of 12 total.
         All models trained on OpenWebText with identical data order and initialization.
         $^{*}$ marks the best ratio per domain; $^{\dagger}$ marks the worst.
         Domains are ordered from most softmax-leaning to most sigmoid-leaning.
         No single configuration dominates across all domains.}
\label{tab:ood}
\resizebox{\textwidth}{!}{%
\begin{tabular}{lrrrrr}
\toprule
\textbf{Domain} & \textbf{$H_{12:0}$} & \textbf{$H_{9:3}$} & \textbf{$H_{6:6}$} & \textbf{$H_{3:9}$} & \textbf{$H_{0:12}$} \\
\midrule
SST-2      & $82.82_{\pm 1.55}^{*}$ & $82.99_{\pm 1.28}$ & $83.72_{\pm 0.85}$ & $85.48_{\pm 1.08}$ & $88.93_{\pm 3.97}^{\dagger}$ \\
Tweets     & $133.65_{\pm 2.18}^{*}$ & $138.60_{\pm 3.85}$ & $140.10_{\pm 4.08}$ & $146.70_{\pm 3.82}$ & $152.92_{\pm 12.00}^{\dagger}$ \\
Reddit     & $38.27_{\pm 0.06}$ & $38.17_{\pm 0.17}$ & $38.14_{\pm 0.24}^{*}$ & $38.38_{\pm 0.24}$ & $39.39_{\pm 0.48}^{\dagger}$ \\
PTB        & $103.13_{\pm 3.70}$ & $100.95_{\pm 2.71}$ & $99.05_{\pm 4.61}^{*}$ & $109.35_{\pm 10.13}$ & $113.09_{\pm 6.64}^{\dagger}$ \\
News       & $24.47_{\pm 0.08}^{*}$ & $24.57_{\pm 0.08}$ & $24.75_{\pm 0.06}$ & $24.79_{\pm 0.27}$ & $24.81_{\pm 0.21}^{\dagger}$ \\
Dialogue   & $30.57_{\pm 0.25}$ & $30.17_{\pm 0.30}^{*}$ & $30.34_{\pm 0.46}$ & $30.36_{\pm 0.25}$ & $30.87_{\pm 0.35}^{\dagger}$ \\
Empathetic & $33.73_{\pm 0.91}^{\dagger}$ & $33.16_{\pm 0.32}$ & $32.72_{\pm 0.82}^{*}$ & $33.48_{\pm 0.44}$ & $33.63_{\pm 0.47}$ \\
LAMBADA    & $52.78_{\pm 0.51}$ & $53.24_{\pm 0.67}$ & $53.18_{\pm 1.08}$ & $53.36_{\pm 1.86}^{\dagger}$ & $52.53_{\pm 1.25}^{*}$ \\
Legal      & $13.49_{\pm 1.36}$ & $12.67_{\pm 1.34}$ & $13.55_{\pm 2.47}^{\dagger}$ & $12.04_{\pm 1.70}$ & $11.69_{\pm 2.00}^{*}$ \\
Code       & $14.40_{\pm 0.79}^{\dagger}$ & $13.58_{\pm 0.64}$ & $13.77_{\pm 1.03}$ & $12.23_{\pm 1.76}$ & $11.67_{\pm 1.06}^{*}$ \\
Novel      & $37.52_{\pm 0.57}^{\dagger}$ & $37.44_{\pm 0.65}$ & $37.33_{\pm 0.63}$ & $37.27_{\pm 0.33}$ & $36.98_{\pm 0.92}^{*}$ \\
Wiki       & $44.68_{\pm 0.90}^{\dagger}$ & $44.31_{\pm 0.76}$ & $43.95_{\pm 1.15}$ & $42.47_{\pm 2.44}$ & $41.25_{\pm 1.89}^{*}$ \\
arXiv      & $104.98_{\pm 2.78}^{\dagger}$ & $100.84_{\pm 3.73}$ & $99.50_{\pm 7.12}$ & $92.44_{\pm 10.57}$ & $76.86_{\pm 2.03}^{*}$ \\
PubMed     & $64.29_{\pm 1.97}^{\dagger}$ & $61.60_{\pm 1.60}$ & $60.51_{\pm 3.75}$ & $55.68_{\pm 5.65}$ & $47.24_{\pm 1.98}^{*}$ \\
Math       & $28.18_{\pm 0.66}^{\dagger}$ & $27.37_{\pm 0.25}$ & $27.57_{\pm 1.36}$ & $25.49_{\pm 3.44}$ & $23.42_{\pm 1.69}^{*}$ \\
\midrule
\# Best ($^{*}$)  & 3 & 1 & 3 & 0 & 8 \\
\# Worst ($^{\dagger}$) & 7 & 0 & 1 & 1 & 6 \\
\bottomrule
\end{tabular}%
}
\end{table*}

This domain-dependence is far larger than Section~\ref{sec:in-domain} would
predict. In-distribution, only $H_{9:3}$ was statistically indistinguishable
from the softmax-only baseline, and the largest degradation across all
ratios was a modest $0.53$ PPL. \textit{Under distribution shift, the gap between
ratios widens by more than an order of magnitude on several domains}---e.g.\
$76.86$ vs.\ $104.98$ PPL on arXiv, a $28$-point spread between $H_{0:12}$
and $H_{12:0}$---and the identity of the better-performing ratio flips
depending on the domain. Ranking the five configurations within each domain,
a pure configuration ($H_{12:0}$ or $H_{0:12}$) is the worst-performing
ratio in 13 of 15 domains, while a mixed ratio is worst in only 2, and in
neither case is the gap from the best configuration statistically
significant (paired $t$-test, $p>0.05$); by contrast, when a pure endpoint
finishes worst, the gap is significant in 11 of 13 cases.\textit{ This pattern is
consistent with the given ratio's influence being suppressed in-distribution
but re-emerging, unevenly across domains, once that constraint is removed.
}

Figure~\ref{fig:pca1_domain} makes this separation visible directly: \textbf{domains
separate cleanly along a single axis of softmax-versus-sigmoid preference},
with the two pure endpoints sitting at opposite ends and $78.3\%$ of the
variance in domain response explained by this one dimension. Because the
identity of the better endpoint flips depending on where a domain falls on
this axis, no fixed ratio is safe to assume in advance.

This axis tracks a structural property of how relevance is distributed
within each domain's documents. SST-2, Tweets, Reddit, and PTB are short,
self-contained units where a handful of tokens carry most of the relevant
signal~\citep{marion2025attention}; arXiv, PubMed, Math, Wiki, and Code
are long, structurally regular documents where relevance is distributed
across many positions—a proof step depends on earlier lemmas, a function
body on earlier bindings~\citep{evaluating-long-range}. Softmax's
sum-to-one allocation suits the former, where suppressing most positions
loses little; sigmoid's independent gating suits the latter, where a
sum-to-one constraint would discard information dense weighting retains.
Dialogue and Empathetic sit closest to the boundary in
Figure~\ref{fig:pca1_domain}, consistent with short turns whose relevance
still accumulates across a conversation. This also accords with the
functional-space view of attention heads as kernel machines
\citep{cheng2024transformers}, where the activation function determines
the class of functions a head can represent.

Two domains depart from this axis-based account. LAMBADA~\cite{paperno2016lambada} sits on the
sigmoid-leaning side of Figure~\ref{fig:pca1_domain} ($-0.80$), yet its
worst-performing ratio is $H_{3:9}$ rather than the pure-sigmoid endpoint
$H_{0:12}$—one of only two domains where a mixed ratio, not a pure one,
finishes last. Empathetic~\cite{rashkin2019towards} is more striking: despite sitting on the
softmax-leaning side ($1.02$), its best ratio is the balanced $H_{6:6}$
and its worst is pure softmax ($H_{12:0}$), the opposite of what the axis
would predict. Both cases involve dialogue-adjacent or long-range-dependency
domains (LAMBADA is explicitly a long-range coreference benchmark), suggesting
the single-axis account captures the dominant structure in domain response
but not domains where short surface form and long-range dependency pull in
different directions simultaneously.

\subsection{Attention Entropy Analysis}
To characterize the internal behavior of MoFA models, we measure mean
attention entropy per head type, averaged across all 15 OOD domains.
Entropy is computed after row-normalizing attention weights, making
sigmoid and softmax heads directly comparable.

\subsubsection{Consecutive-Ratio Comparisons}
\label{sec:consecutive-ratio}
To test whether the reorganization between softmax and sigmoid heads is
gradual or threshold-like, we compare domain-averaged entropy between each
pair of adjacent ratios via paired $t$-tests across the five seeds
(Table~\ref{tab:entropy_consecutive}). Sigmoid head entropy decreases
significantly at every step ($p<0.05$ for all three transitions), but the
magnitude of the decrease grows sharply toward the pure-sigmoid endpoint:
$-0.05$ ($H_{9:3}\to H_{6:6}$), $-0.11$ ($H_{6:6}\to H_{3:9}$), and $-0.36$
($H_{3:9}\to H_{0:12}$, $p<10^{-4}$)---an accelerating decline.
Softmax head entropy shows the opposite profile: 
the first two transitions are not statistically significant
($p=0.52$, $p=0.44$), and only the $H_{6:6}\to H_{3:9}$ step reaches
significance ($p=0.035$), consistent with softmax heads retaining a stable
role across most configurations and reorganizing only once sigmoid heads
constitute the majority.

\begin{table}[htbp]
\centering
\caption{Paired $t$-tests (across 5 seeds) comparing domain-averaged head
         entropy between consecutive ratios. $\Delta$ is the change from the
         first to the second ratio; $^{*}$ marks $p<0.05$.}
\label{tab:entropy_consecutive}
\begin{tabular}{lcccc}
\toprule
& \multicolumn{2}{c}{\textbf{Softmax heads}} & \multicolumn{2}{c}{\textbf{Sigmoid heads}} \\
\textbf{Transition} & $\Delta$ & $p$ & $\Delta$ & $p$ \\
\midrule
$H_{12:0} \to H_{9:3}$ & $-0.013$ & $0.522$ & --- & --- \\
$H_{9:3} \to H_{6:6}$  & $+0.031$ & $0.439$ & $-0.051$ & $0.047^{*}$ \\
$H_{6:6} \to H_{3:9}$  & $-0.146$ & $0.035^{*}$ & $-0.109$ & $0.003^{*}$ \\
$H_{3:9} \to H_{0:12}$ & --- & --- & $-0.361$ & $<0.001^{*}$ \\
\bottomrule
\end{tabular}
\end{table}

\subsubsection{Where This Occurs in the Network}
\begin{figure}[htbp]
    \centering
    \begin{minipage}{0.48\textwidth}
        \centering
        \includegraphics[width=\textwidth]{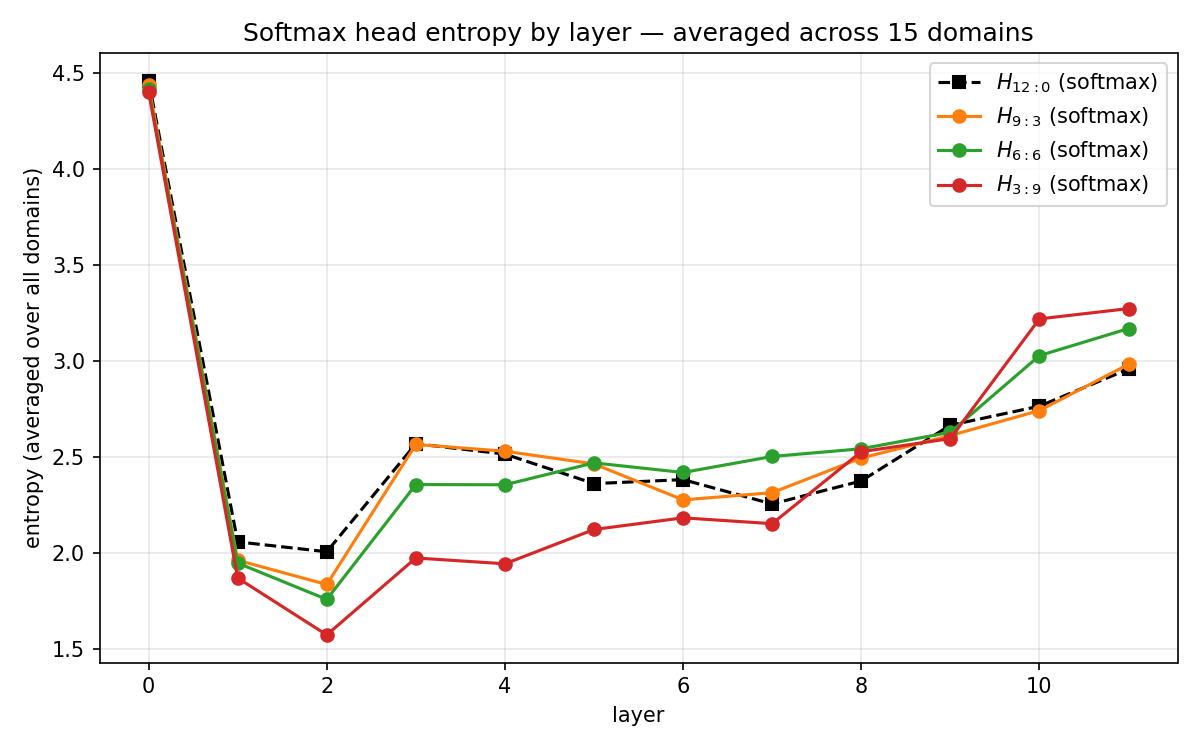}
    \end{minipage}
    \hfill
    \begin{minipage}{0.48\textwidth}
        \centering
        \includegraphics[width=\textwidth]{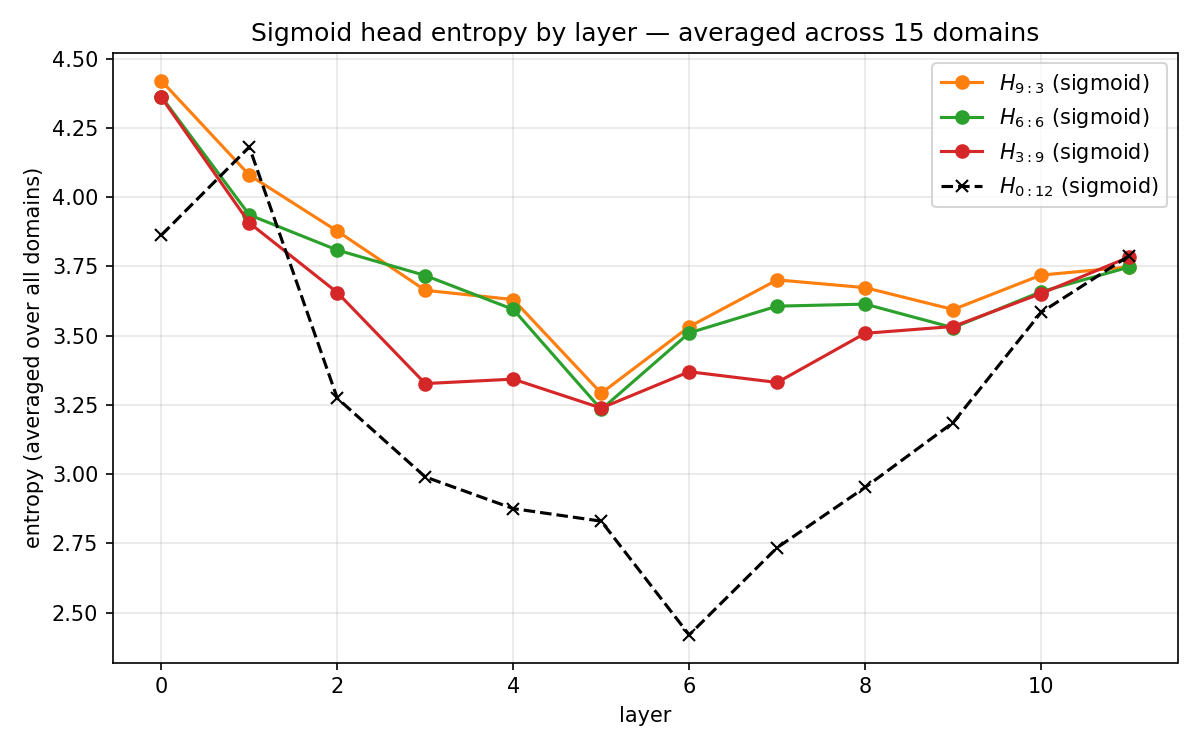}
    \end{minipage}
    \caption{Per-layer attention entropy, averaged across 15 OOD domains, for softmax heads (left) and sigmoid heads (right) under each ratio. Dashed black lines with square markers denote the pure endpoints ($H_{12:0}$ left, $H_{0:12}$ right); solid colored lines denote the mixed ratios.}
    \label{fig:division_of_labor}
\end{figure}

Figure~\ref{fig:division_of_labor} localizes the asymmetry from 
Section~\ref{sec:consecutive-ratio} within the network. 
We report these layer-level patterns as descriptive observations.

Softmax head entropy tracks a mostly stable curve regardless of how many sigmoid heads are present, with a mild divergence at layers 2--4 (Fig.~\ref{fig:division_of_labor}, left). 
Sigmoid head entropy remains within a narrow, elevated band across the three mixed ratios and drops sharply only in $H_{0:12}$ (Fig.~\ref{fig:division_of_labor}, right), localizing the accelerating decline from Section~\ref{sec:consecutive-ratio} to a mid-network trough that deepens specifically once no softmax heads remain.

% \begin{figure}[htbp]
%     \centering
%     \begin{minipage}{0.48\textwidth}
%         \centering
%         \includegraphics[width=\textwidth]{images/division_of_labor_compensation_sfx.png}
%     \end{minipage}
%     \hfill
%     \begin{minipage}{0.48\textwidth}
%         \centering
%         \includegraphics[width=\textwidth]{images/division_of_labor_compensation_sig.png}
%     \end{minipage}
%     \caption{Compensation effect at each activation's strongest-response layer. \textbf{Left:} softmax entropy at layer 2 vs.\ number of sigmoid heads present. \textbf{Right:} sigmoid entropy at layer 6 vs.\ number of remaining softmax heads. Error bars: std over domain$\times$seed.}
%     \label{fig:compensation}
% \end{figure}

% At each activation's strongest-response layer, the same asymmetry sharpens:
% softmax entropy at layer 2 declines gradually as sigmoid heads are added,
% while sigmoid entropy at layer 6 barely shifts across $H_{9:3}$--$H_{3:9}$ 
% and drops sharply only in $H_{0:12}$---the same accelerating pattern observed in
% Section~\ref{sec:consecutive-ratio}.

\section{Discussion}
\label{sec:discussion}
Two aspects of our results go beyond the suppression-and-reemergence pattern itself and are worth drawing out.

The \textit{form} of reemergence is itself informative: it is not uniform variance but a single, interpretable axis (Figure~\ref{fig:pca1_domain}). If activation choice functioned merely as unstructured noise unmasked by the removal of training pressure, we would expect the ranking of ratios to vary idiosyncratically across domains.
Instead, 78.3\% of domain response collapses onto one dimension tracking document-level relevance structure---local vs.\ distributed. This suggests the given prior does not simply ``reawaken'' under weak constraints; it reasserts a \textit{specific}, architecturally-determined preference (softmax's competitive allocation vs.\ sigmoid's independent gating) that was already latent in the mechanism, and which the OOD domain's structure either rewards or penalizes. The two domains that depart from this axis (LAMBADA, Empathetic; Section~\ref{subsubsec:domain_perfomance}) indicate that this dimension is dominant but not exhaustive---a more granular account of domain structure is a natural next step.
The role of distribution shift, on this reading, is not to inject variance but to remove the one corpus statistic (Section~\ref{sec:in-domain}) that had been overriding this latent preference.

A related asymmetry links our head-level and domain-level results.
At the head level, sigmoid heads reorganize continuously as their share of the network grows, while softmax heads hold a stable role until they stop being the majority (Table~\ref{tab:entropy_consecutive}).
At the domain level, we see the same pattern: pure endpoints are disproportionately likely to be the \textit{worst}-performing configuration (13/15 domains), while mixed ratios finish worst in only two. 
Both point to the same underlying effect—heads interact with their neighbors' activation type, so that removing one type entirely changes how the other behaves, even though no head's own mechanism has changed.
We do not have a mechanistic explanation for why a minority presence stabilizes the majority's role, and we leave this as a direction for follow-up work.

\paragraph{Limitations.} Our experiments are limited to a single model scale (124M parameters), a single pretraining corpus (OpenWebText), and a single pair of activation functions (softmax and sigmoid); it remains open whether the same suppression-and-reemergence pattern holds at larger scales, other pretraining distributions, or other activation combinations. Our layer-level analysis is descriptive without statistical rigor.

\section{Conclusion}
\label{sec:conclusion}
We asked whether the attention function given to a network can serve as an intrinsic inductive bias, and how its influence changes as the network generalizes to novel environments. In-distribution, this influence is largely suppressed; under distribution shift, it reasserts itself sharply and unevenly, tracking a single axis of domain structure and reorganizing attention at the head level. Treating activation choice as such an intrinsic prior offers a productive lens on how transformers generalize.

\section{Broader Impact}
\label{sec:broader-impact}

This work studies a specific architectural design choice—the mix of activation functions across attention heads—and its effect on language modeling perplexity under distribution shift. 
We treat activation choice as a given prior, drawing on core-knowledge accounts of inductive bias to motivate both MoFA's design and its evaluation under generalization from limited experience.

\section{Reproducibility}
\label{sec:reproducibility}

All experiments use publicly available data, and the code is available at
\url{https://github.com/DoGy-Kang/MoFA_public}. The 124M-parameter
GPT-2 architecture, training procedure, and optimizer settings are fully
specified in Section~4.1; all five ratio configurations share identical
data order, weight initialization, and hyperparameters across five fixed
seeds, differing only in the sigmoid head
count $H_{\text{sig}}$. The custom BPE tokenizer is trained from scratch
on OpenWebText \citep{gokaslan2019openwebtext} with a fixed vocabulary of
50{,}000 tokens. All out-of-distribution evaluation datasets are
publicly available and used only for zero-shot perplexity measurement
(Appendix~\ref{app:datasets}). Statistical tests (paired $t$-tests)
are computed across the five seeds using standard implementations, with
exact $p$-values reported in Tables~1--3.

\newpage

\bibliographystyle{unsrtnat}
\bibliography{references}

\appendix
\section*{Appendix}
\section{Dataset Details}
\label{app:datasets}

Table~\ref{tab:datasets} lists the HuggingFace dataset identifier, split, and
original source publication for each of the 15 out-of-distribution domains
evaluated in Section~4.3. All datasets are evaluated zero-shot with no
fine-tuning; text is tokenized with the custom 50k-vocabulary BPE tokenizer
trained on OpenWebText (Section~4.1) and chunked into blocks following the
grouping strategy described in the released evaluation code.

\begin{table}[h]
\centering
\caption{Dataset sources for the 15 OOD evaluation domains.}
\label{tab:datasets}
\small
\begin{tabular}{llll}
\toprule
\textbf{Domain} & \textbf{HF Identifier} & \textbf{Split} & \textbf{Source} \\
\midrule
Code       & \texttt{code\_search\_net} (python)      & test           & \citet{husain2019codesearchnet} \\
Novel      & \texttt{bookcorpus}                      & train[:5\%]    & \citet{zhu2015aligning} \\
News       & \texttt{cnn\_dailymail} (3.0.0)           & test           & \citet{see2017get} \\
PubMed     & \texttt{ccdv/pubmed-summarization}        & test           & \citet{cohan2018discourse} \\
arXiv      & \texttt{ccdv/arxiv-summarization}          & test           & \citet{cohan2018discourse} \\
Wiki       & \texttt{wikitext} (103-v1)                & test           & \citet{merity2016pointer} \\
Reddit     & \texttt{reddit\_tifu} (long)              & train[:5\%]    & \citet{kim2019abstractive} \\
Legal      & \texttt{billsum}                          & test           & \citet{kornilova2019billsum} \\
Dialogue   & \texttt{roskoN/dailydialog}                & test           & \citet{li2017dailydialog} \\
Empathetic & \texttt{facebook/empathetic\_dialogues}    & test           & \citet{rashkin2019towards} \\
Tweets     & \texttt{cardiffnlp/tweet\_eval} (emotion)  & test           & \citet{barbieri2020tweeteval} \\
SST-2      & \texttt{stanfordnlp/sst2}                  & validation     & \citet{socher2013recursive} \\
LAMBADA    & \texttt{EleutherAI/lambada\_openai}        & test           & \citet{paperno2016lambada} \\
Math       & \texttt{gsm8k} (main)                      & test           & \citet{cobbe2021training} \\
PTB        & \texttt{ptb\_text\_only}                   & test           & \citet{marcus1993building} \\
\bottomrule
\end{tabular}
\end{table}

Novel and Reddit use \texttt{train[:5\%]} splits because their source
datasets do not provide a dedicated test split; all other domains use the
canonical test (or validation, for SST-2) split. arXiv and PubMed share
the same underlying summarization dataset construction [Cohan et al., 2018]
but are drawn from disjoint document collections (arXiv preprints vs.\
PubMed abstracts).

\end{document}